\pdfoutput=1

\documentclass[11pt]{article}

\usepackage{EACL2023}

\usepackage{times}
\usepackage{latexsym}

\usepackage[T1]{fontenc}

\usepackage[utf8]{inputenc}

\usepackage{microtype}

\usepackage{inconsolata}

\usepackage{booktabs}
\usepackage{amsmath}
\usepackage{amsthm}
\newtheorem{theorem}{Theorem}
\usepackage{graphicx}
\usepackage{enumitem}
\usepackage{soul}
\usepackage{algorithm}
\usepackage[noend]{algpseudocode}

\title{EDU-level Extractive Summarization with Varying Summary Lengths}

\author{
Yuping Wu, Ching-Hsun Tseng, Jiayu Shang, Shengzhong Mao, \\
\bf{Goran Nenadic, Xiao-Jun Zeng\thanks{\, Corresponding author.}} \\
Department of Computer Science, University of Manchester \\
\texttt{\{yuping.wu-2, ching-hsun.tseng, jiayu.shang,} \\ \texttt{shengzhong.mao\}@postgrad.manchester.ac.uk}\\
\texttt{gnenadic, x.zeng@manchester.ac.uk}
      }

\begin{document}
\maketitle

\begin{abstract}

Extractive models usually formulate text summarization as extracting fixed top-$k$ salient sentences from the document as a summary. Few works exploited extracting finer-grained Elementary Discourse Unit (EDU) with little analysis and justification for the extractive unit selection. Further, the selection strategy of the fixed top-$k$ salient sentences fits the summarization need poorly, as the number of salient sentences in different documents varies and therefore a common or best $k$ does not exist in reality. To fill these gaps, this paper first conducts the comparison analysis of oracle summaries based on EDUs and sentences, which provides evidence from both theoretical and experimental perspectives to justify and quantify that EDUs make summaries with higher automatic evaluation scores than sentences. Then, considering this merit of EDUs, this paper further proposes an \textbf{EDU}-level extractive model with \textbf{V}arying summary \textbf{L}engths (EDU-VL\footnote{\url {https://github.com/yuping-wu/EDU-VL}}) and develops the corresponding learning algorithm. EDU-VL learns to encode and predict probabilities of EDUs in the document, generate multiple candidate summaries with varying lengths based on various $k$ values, and encode and score candidate summaries, in an end-to-end training manner. Finally, EDU-VL is experimented on single and multi-document benchmark datasets and shows improved performances on ROUGE scores in comparison with state-of-the-art extractive models, and further human evaluation suggests that EDU-constituent summaries maintain good grammaticality and readability.

\end{abstract}

\section{Introduction}

Automatic text summarization aims at aggregating information in long document(s) into a shorter piece of text while keeping important information. Extractive summarization and abstractive summarization are two categories of it. This paper focuses only on the extractive task which formulates summarization as identifying salient textual segments in document \citep{Lunh1958TheAbstracts}. Under the supervised learning framework, this task is further formulated as a label classification task, i.e., encoding textual segments and predicting labels on the encoded vectors. Recent state-of-the-art models (\citealp{liu-lapata-2019-text}; \citealp{zhong-etal-2020-extractive}; \citealp{liu-etal-2021-hetformer}; \citealp{ruan-etal-2022-histruct}) on this task tend to be Transformer-based since BERT \citep{devlin-etal-2019-bert} shows significantly better performance than RNN on most natural language understanding tasks.

\begin{table}[t]
    \centering
    \begin{tabular}{p{0.9\linewidth}}
    \hline
        \textit{\textbf{Document:}} (...) [\textcolor{blue}{The second audio,}] [taken from dash cam video from inside a patrol car,] [\textcolor{blue}{captures a phone call between Slager and someone}] [\textcolor{blue}{CNN believes}] [\textcolor{blue}{is his wife.}] (...) \\
        \hline
        \textit{\textbf{Reference Summary: }} \textcolor{blue}{The second audio captures a phone call between Slager and someone CNN believes is his wife.} \\
        \hline
    \end{tabular}
    \caption{Example to demonstrate redundant information in sentence. Content within [] indicates an EDU.}
    \label{tab:edu example}
\end{table}

Most existing works extract sentences from the document and some works further \citep{xu-durrett-2019-neural} propose post-processing steps to prune the generated summary. The only exception is the few works (\citealp{liu-chen-2019-exploiting}; \citealp{huang-kurohashi-2021-extractive}), which extract finer-grained textual segments, i.e., discourse-level text or EDU, with little justification. The intuition is that a sentence consisting of multiple clauses is inevitable to contain less important information. As demonstrated in Table \ref{tab:edu example}, partially removing a clause in the sentence is conducive to generating a summary. Certainly, such an intuitive explanation does not provide enough evidence and support to justify the use of finer-grained textual segments such as EDU to substitute sentences. Considering such a gap in existing research, the first main motivation of this paper is to propose and conduct the comparison analysis between sentences and EDUs to disclose and justify whether using EDU is a theoretically advanced and application-advantaged extractive unit.


When selecting textual segments, the top-$k$ strategy with $k$ fixed for all documents is dominant in deciding the length of the generated summary. Some works (\citealp{zhong-etal-2020-extractive}; \citealp{chen-etal-2021-sgsum}) manage to output summaries with different lengths, i.e., various numbers of extracted segments, via formulating the problem as deriving a subset of sentences from the combination of top-$k$ sentences. Due to the foreseeing explosion of the combination of sentences to form subsets, these approaches are limited to generating summaries with relatively small values of $k$. To overcome such a weakness, the second main motivation of this paper is to propose and develop an approach allowing varying lengths for extractive summarization without explicit limitation on the maximum value of $k$, i.e., the maximum length.

Following the above motivations, the comparison analysis between EDUs and sentences ascertains that EDU is a better text unit for the extractive task because EDU-level summaries achieve higher automatic evaluation scores than sentence-level summaries. This conclusion is justified from two perspectives. Theoretically, a formal theorem about this conclusion could be derived from the property that EDU is essentially part of a sentence. Experimentally, results of comprehensive analysis about oracle summaries of five datasets further quantify this conclusion, i.e., how much the ROUGE scores of EDU-level oracle summary are higher than sentence-level oracle summary.

Based on the aforementioned conclusion and foundation, this paper further proposes and develops an EDU-level extractive model and algorithm, which generates summaries with varying lengths, i.e., EDU-VL. We extend Transformer-based pre-trained language model with an extra classification layer to encode EDUs in a document and predict the corresponding probabilities. Multiple $k$ values are provided to the model to generate a set of candidate summaries under the flexible top-$k$ strategy for the document. Multiple Transformer encoder layers encode the full document and candidate summaries individually. Finally, a similarity score with the encoded document is calculated for each candidate summary and the one with the highest score is the final output of EDU-VL.

Experiments are conducted on five benchmark datasets from different domains and with various writing styles. The experimental results suggest that EDU-VL achieves better performance than all state-of-the-art extractive baselines on single-document summarization datasets CNN/DailyMail, XSum, Reddit, and WikiHow, in terms of three ROUGE metrics. With direct comparison to the multi-document model, EDU-VL still achieves comparable performance on the multi-document summarization dataset Multi-News. Human evaluation is further carried for the summaries generated by EDU-VL to assess the syntax structure of EDU-constituent summaries. The results provide evidence for the good grammaticality and readability of EDU-constituent summaries and therefore justify the applicability.

The contributions of this paper are threefold:
\begin{enumerate}[label=\arabic*)]
    \item We justify and quantify that EDU-level achieves higher automatic evaluation scores than sentence-level oracle summary from both theoretical and experimental perspectives, indicating that setting EDU as the extractive text unit is exploitable and superior in applications.
    \item We propose a varying summary lengths-enabled extractive model with EDU-level text unit. Such a model and its learning algorithm encodes EDUs in a document and outputs a summary with varying length by making $k$ in the top-$k$ extraction strategy varying.
    \item Our proposed model achieves superior performance on four single-document summarization datasets on three ROUGE metrics. Human evaluations show that the generated EDU-constituent summaries maintain good grammaticality and readability.
\end{enumerate}

\section{Related Work}

\subsection{Neural Extractive Summarization}
The extractive text summarization task aims at extracting salient textual segments from the original document(s) as a summary. A tendency observed among extractive neural models is that the architecture changes from RNN (\citealp{Nallapati2017SummaRuNNer:Documents}; \citealp{xu-durrett-2019-neural}) to Transformer-based models, e.g., BERT (\citealp{zhang-etal-2019-hibert}; \citealp{liu-lapata-2019-text}) and Longformer (\citealp{liu-etal-2021-hetformer}; \citealp{ruan-etal-2022-histruct}). GNN also gained extensive attention in recent years and is usually stacked after an RNN (\citealp{wang-etal-2020-heterogeneous}; \citealp{jing-etal-2021-multiplex}) or Transformer-based encoder (\citealp{cui-etal-2020-enhancing}; \citealp{kwon-etal-2021-considering}) to supplement graph-based features. Some research works integrated neural networks with reinforcement learning (\citealp{dong-etal-2018-banditsum}; \citealp{gu-etal-2022-memsum}) or unsupervised learning frameworks \citep{liang-etal-2021-improving}. In general, it can be said that taking a pre-trained Transformer-based language model as the starting point to encode textual segments in a document is currently the state-of-the-art approach among neural extractive models. Therefore, the Transformer-based models, i.e., RoBERTa \citep{Liu2019RoBERTa:Approach} and BART \citep{lewis-etal-2020-bart}, are used as the basic building blocks in this paper.


\subsection{Sub-sentential Extractive Summarization}
Most previous works about the extractive task focused on generating sentence-level summaries, though some of them (\citealp{xiao-etal-2020-really}; \citealp{cho-etal-2020-better}; \citealp{Ernst2022Proposition-LevelSummarization}) utilized sub-sentential features. Early works by \citet{Marcu1999DiscourseText}; \citet{AlonsoiAlemany2003IntegratingSummarization}; \citet{yoshida-etal-2014-dependency}; \citet{li-etal-2016-role} exploited extracting discourse-level textual segments as the summary but those approaches were tested on small datasets. More recent works by  \citet{liu-chen-2019-exploiting}; \citet{xu-etal-2020-discourse}; \citet{huang-kurohashi-2021-extractive} were evaluated on relatively larger datasets. However, whether the discourse-level textual segments are a better alternative than sentences as the extractive text unit was not justified in those works. To fill this gap, we provide justification for this research question from both theoretical and experimental perspectives in this paper.

\subsection{Flexible Extractive Summarization}
Extractive summarization task is usually formulated as extracting the top-$k$ number of salient textual segments from a document. The fixed $k$ value for all documents results in the lack of variety in the length of the generated summary. Few works (\citealp{jia-etal-2020-neural}; \citealp{zhong-etal-2020-extractive}; \citealp{chen-etal-2021-sgsum}) managed to output summaries with varying lengths. However, either it requires extra effort for hyper-parameter searching on validation dataset to find a valid threshold, or formulating the problem as selecting a subset of top-$k$ sentences makes the variety of lengths limited to small lengths due to the explosive nature of combination. In this paper, we propose a model with varying $k$ values but without explicit limitation on the length or the need to do hyper-parameter searching.

\section{Oracle Analysis of EDUs and Sentences}
Oracle analysis refers to the analysis of oracle summary whose definition is stated in Section \ref{definition: os}. We conducted oracle analysis from both theoretical and experimental perspectives to justify and quantify that discourse-level summary achieves higher scores on automatic evaluation metrics than sentence-level summary.

\subsection{Theoretical Formulation}
Elementary Discourse Unit (EDU), the discourse-level textual segment in this paper, refers to the terminal node in the Rhetorical Structure Theory (RST) \citep{Mann1988RhetoricalOrganization} tree which describes the discourse structure of a piece of text. EDUs are non-overlapping and adjacent text spans in the piece of text and a single EDU is essentially a segment of a complete sentence, i.e., the sentence itself or a clause in the sentence \citep{zeldes-etal-2019-disrpt}. Namely, a sentence can always be expressed with multiple EDUs, i.e., for the \(s\)-th sentence in a document, there is \(sent_s=[edu_{s_1},\dots,edu_{s_m}]\). Consequently, a one-way property from sentence to EDU regarding expressiveness is derived. 

\paragraph{Expressiveness Property}\label{property: expressiveness} For any given subset of sentences in a document, i.e., $[sent_i, \dots, sent_j, \dots, sent_k]$, there is always a subset of EDUs in the document, i.e., $[edu_{i_1},\dots,edu_{i_m}, \dots, edu_{j_1}, \dots, edu_{j_m},\dots, \\ edu_{k_1},\dots,edu_{k_m}]$, having identical content.


\paragraph{Oracle Summary}\label{definition: os} The set of salient textual segments that have greedily the highest ROUGE score(s) with the reference summary is the oracle summary for a document. It signifies the upper bound of performance that an extractive summarization model could achieve on ROUGE metrics.

Denote the sentence-level oracle summary as \(\mathcal{OS}_{sent}\) and the EDU-level oracle summary as \(\mathcal{OS}_{edu}\). Based on the aforementioned property and definition, Theorem \ref{oracle theorem} can be derived and its detailed proof is provided below.

\begin{theorem} \label{oracle theorem}
Given a document $\mathcal{D}$ and its reference summary $\mathcal{R}$, for any derived $\mathcal{OS}_{sent}$, there is always an $\mathcal{OS}_{edu}$ having ROUGE\textsubscript{F\textsubscript{1}}$(\mathcal{R}, \mathcal{OS}_{edu}) \geq$ ROUGE\textsubscript{F\textsubscript{1}}$(\mathcal{R}, \mathcal{OS}_{sent})$.
\end{theorem}

\begin{proof}
For ROUGE-N, let $f_n$ be a function that generates the set of n-grams for the string $s$ and $g$ be a function that calculates the number of overlapping elements between two sets $x$ and $y$, i.e., \\
\centerline{$f_n(s)=n$-$gram(s)$,} \\
\centerline{$g(x,y)=match(x,y)$.} \\
\\
The recall and precision formulas of the ROUGE-N metric between the reference summary $\mathcal{R}$ and sentence-level oracle summary $\mathcal{OS}_{sent}$ are \\
\centerline{R-N\textsubscript{recall, $\mathcal{OS}_{sent}$} = $\frac{g(f_n(\mathcal{R}), f_n(\mathcal{OS}_{sent}))}{\vert f_n(\mathcal{R}) \vert}$,} \\
\centerline{R-N\textsubscript{precision, $\mathcal{OS}_{sent}$} = $\frac{g(f_n(\mathcal{R}), f_n(\mathcal{OS}_{sent}))}{\vert f_n(\mathcal{OS}_{sent}) \vert}$.} \\
\\
There is always an EDU-level summary $\mathcal{S}_{edu}$ having $\mathcal{S}_{edu}$ = $\mathcal{OS}_{sent}$. Let $\mathcal{S}_{edu}^{sub}$ be the subset of EDUs in $\mathcal{S}_{edu}$ having equivalent number of overlapping n-grams as $\mathcal{S}_{edu}$, i.e., \\
\\
\centerline{$\mathcal{S}_{edu}^{sub} \subseteq \mathcal{S}_{edu} = \mathcal{OS}_{sent}$} \\
and \\
\centerline{$g(f_n(\mathcal{R}), f_n(\mathcal{S}_{edu}^{sub}))=g(f_n(\mathcal{R}), f_n(\mathcal{OS}_{sent}))$.} \\
\\
The number of words in $\mathcal{S}_{edu}^{sub}$ is smaller than or equal to the number of words in $\mathcal{OS}_{sent}$, i.e.,  \\
\\
\centerline{$\vert\mathcal{S}_{edu}^{sub}\vert \le \vert\mathcal{OS}_{sent}\vert$,} \\
\\
and consequently, the number of n-grams is correspondingly smaller or equal, i.e.,  \\
\\
\centerline{$\vert f_n(\mathcal{S}_{edu}^{sub}) \vert \le \vert f_n(\mathcal{OS}_{sent}) \vert$.} \\
\\
Therefore, the precision score for $\mathcal{S}_{edu}^{sub}$ is larger than or equal to $\mathcal{OS}_{sent}$ and their recall scores are the same, i.e., \\
\\
\centerline{R-N\textsubscript{precision, $\mathcal{S}_{edu}^{sub}$} = $\frac{g(f_n(\mathcal{R}), f_n(\mathcal{S}_{edu}^{sub}))}{\vert f_n(\mathcal{S}_{edu}^{sub}) \vert}$} \\
\centerline{$\ge$} \\
\centerline{R-N\textsubscript{precision, $\mathcal{OS}_{sent}$} =$\frac{g(f_n(\mathcal{R}), f_n(\mathcal{OS}_{sent}))}{\vert f_n(\mathcal{OS}_{sent}) \vert}$} \\
and \\
\centerline{R-N\textsubscript{recall, $\mathcal{S}_{edu}^{sub}$} = R-N\textsubscript{recall, $\mathcal{OS}_{sent}$}} \\
\\
Therefore, the EDU-level subset of $\mathcal{OS}_{sent}$, i.e., $\mathcal{S}_{edu}^{sub}$, is found to have higher or equal F1-scores on ROUGE-N metrics than $\mathcal{OS}_{sent}$, i.e., \\
\\
\centerline{R-N\textsubscript{F\textsubscript{1},$\mathcal{S}_{edu}^{sub}$} $\ge$ R-N\textsubscript{F\textsubscript{1},$\mathcal{OS}_{sent}$}} \\
\\
That is to say, it is guaranteed to have an EDU-level summary having higher or equal R-N scores than $\mathcal{OS}_{sent}$. By taking this $\mathcal{S}_{edu}^{sub}$ as $\mathcal{OS}_{edu}$, we have R-N\textsubscript{F\textsubscript{1},$\mathcal{OS}_{edu}$} $\ge$ R-N\textsubscript{F\textsubscript{1},$\mathcal{OS}_{sent}$}. \\
\\
A similar proof process can be conducted on ROUGE-L. Therefore, for any $\mathcal{OS}_{sent}$, there is always an $\mathcal{OS}_{edu}$ having ROUGE\textsubscript{F\textsubscript{1}}$(\mathcal{R}, \mathcal{OS}_{edu}) \geq$ ROUGE\textsubscript{F\textsubscript{1}}$(\mathcal{R}, \mathcal{OS}_{sent})$.

\end{proof}

\subsection{Empirical Justification}
\label{section:empirial justification}
\begin{table}
    \centering
    \begin{tabular}{lccc}
        \hline
        Text Unit & R-1 & R-2 & R-L \\
        \hline
        \multicolumn{4}{c}{\textbf{CNN/DailyMail}} \\
        \hline
        Sentence & 53.33 & 31.09 & 49.67 \\
        EDU & \textbf{61.02} & \textbf{37.16} & \textbf{58.63} \\
        \hline
        \multicolumn{4}{c}{\textbf{XSum}} \\
        \hline
        Sentence & 29.13 & 8.70 & 22.32 \\
        EDU & \textbf{36.07} & \textbf{11.74} & \textbf{30.95} \\
        \hline
        \multicolumn{4}{c}{\textbf{WikiHow}} \\
        \hline
        Sentence & 37.98 & 13.76 & 35.18 \\
        EDU & \textbf{44.28} & \textbf{17.94} & \textbf{42.56} \\
        \hline
        \multicolumn{4}{c}{\textbf{Reddit}} \\
        \hline
        Sentence & 30.58 & 10.95 & 24.57 \\
        EDU & \textbf{40.62} & \textbf{16.01} & \textbf{35.95} \\
        \hline
        \multicolumn{4}{c}{\textbf{Multi-News}} \\
        \hline
        Sentence & 49.65 & 22.20 & 44.99 \\
        EDU & \textbf{51.35} & \textbf{23.99} & \textbf{48.70} \\
        \hline
    \end{tabular}
    \caption{ROUGE F1-scores of sentence-level and EDU-level oracle summaries on training datasets.}
    \label{table:oracle analysis}
\end{table}


Five datasets from different domains were analyzed from the experimental perspective and experimental settings are listed in Appendix \ref{appendix: os length}. Table \ref{table:oracle analysis} presents the ROUGE scores of \(\mathcal{OS}_{sent}\) and \(\mathcal{OS}_{edu}\) on training datasets. \(\mathcal{OS}_{edu}\) gains significantly higher ROUGE scores on all datasets. Larger improvements are observed on ROUGE-1 (6.3-10.04) and ROUGE-L (7.38-11.38) on the majority of datasets, and improvement on ROUGE-2 is smaller but there is still an increase.

\begin{figure}
    \centering
    \includegraphics[scale=0.42]{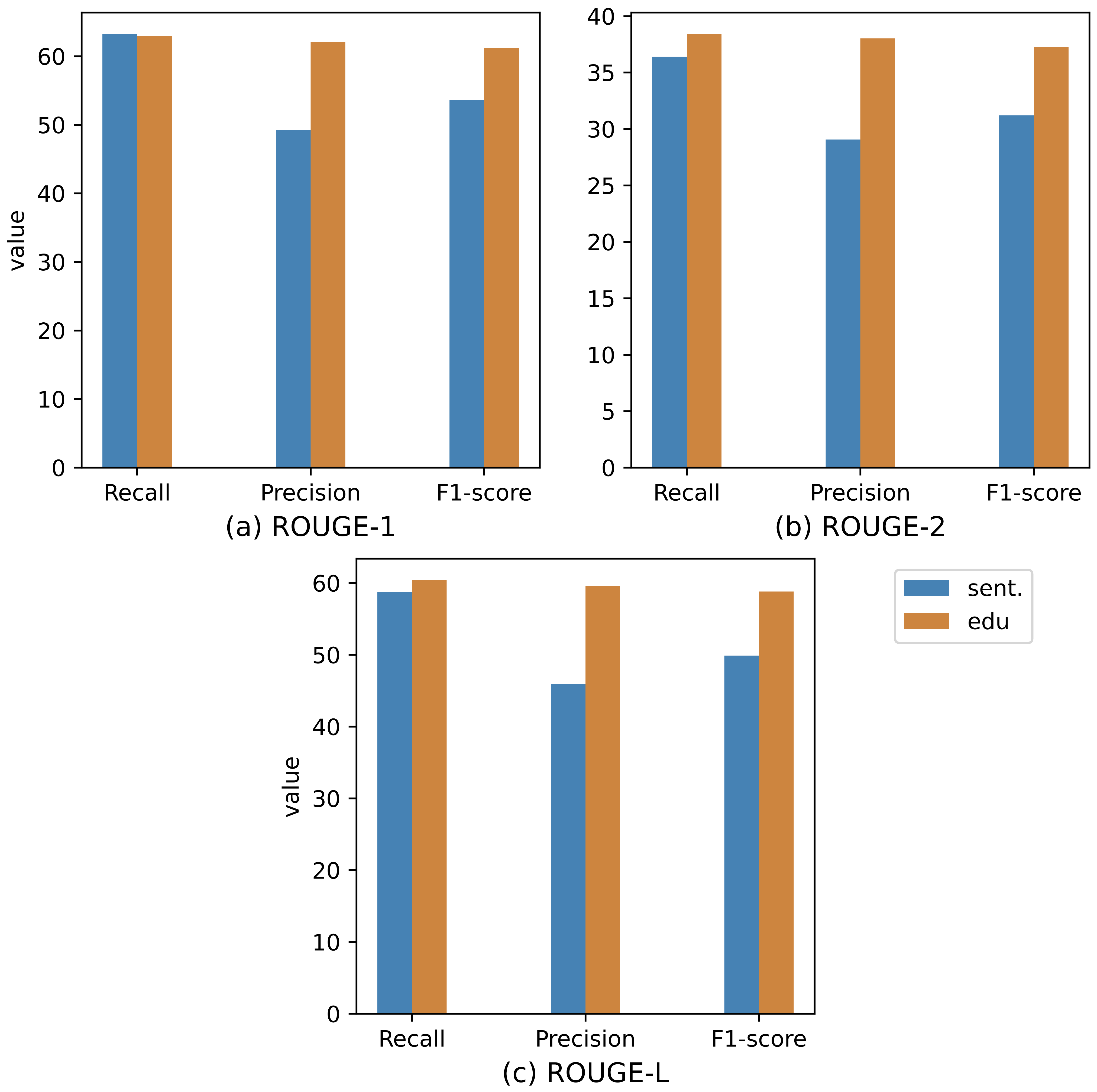}
    \caption{Breakdown ROUGE scores of sentence/EDU-level oracle summaries on CNN/DM training dataset.}
    \label{fig:cnndm_os_plot}
\end{figure}

Figure \ref{fig:cnndm_os_plot} shows the comparison of breakdown ROUGE scores between two text units on the CNN/DailyMail training dataset and details about other datasets could be found in Appendix \ref{appendix: breakdown rouge}. Recall scores on all three metrics are approximately equal between the two text units, suggesting that the amount of salient information in both is equal. However, precision scores are observed with a significantly higher value on $\mathcal{OS}_{edu}$, suggesting the length of $\mathcal{OS}_{edu}$ is smaller.

The experimental results quantify the potential gains that EDU-level oracle summary could achieve on five datasets and the breakdown scores indicate that EDU-level oracle summary is less redundant than sentence-level oracle summary.

\section{EDU-level Extractive Model with Varying Summary Lengths}
\subsection{Problem Formulation}

Suppose a document \(\mathcal{D}\) consists of \(m\) EDUs, i.e., \(\mathcal{D}=[edu_1,\dots, edu_m]\), the \(i\)-th EDU consists of \(n_i\) words, i.e., \(edu_i=[w_{i1},\dots, w_{in_i}]\), and the reference summary wrote by human is denoted as \(\mathcal{R}\). The set of ground truth labels for each EDU could be derived from $\mathcal{R}$, i.e., \(L=[l_1,\dots,l_{m}]\), via a greedy algorithm as previous works did. Our proposed model aims to generate a summary via selecting one summary from the set of candidate summaries $\mathcal{C}$ where \(\mathcal{C}=[cand_1,\dots,cand_c]\) and $cand_j$ consists of EDUs with top-$k_j$ probabilities that are also predicted by the proposed model. 

\subsection{Model}

\begin{figure*}[ht]
    \centering
    \includegraphics[scale=0.12]{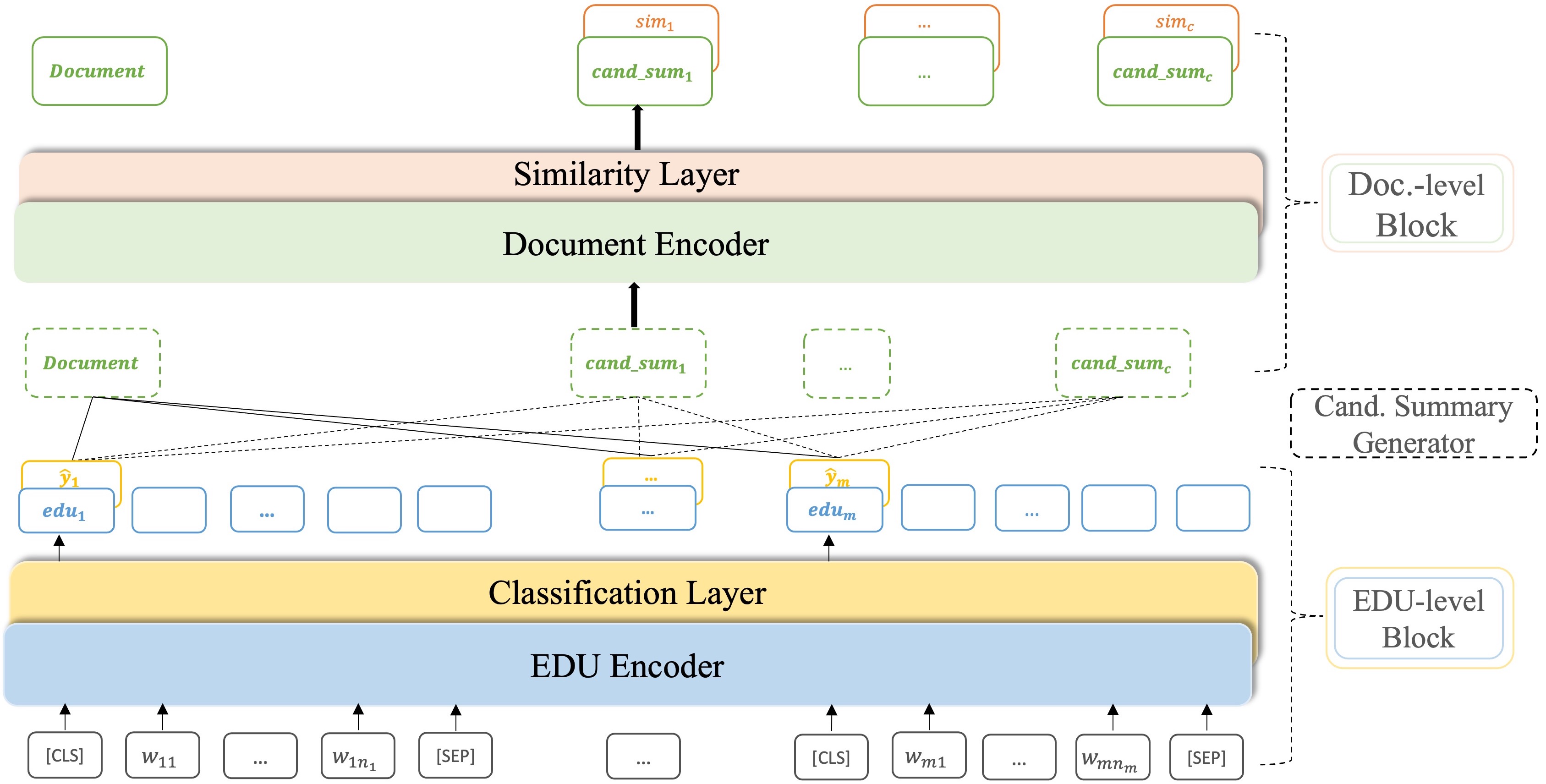}
    \caption{Model architecture. The EDU-level block encodes and predicts a probability value for each EDU in the input document; The candidate summary generator generates a set of candidate summaries based on the predicted probability values; The document-level block encodes the whole document and candidate summaries and generates similarity values between them. The final output is the candidate summary with the highest similarity score.}
    \label{fig:architecture}
\end{figure*}

Figure \ref{fig:architecture} illustrates the architecture of our proposed model. From bottom to top, firstly, the EDU-level block generates a representation vector and probability for each EDU in a document. Secondly, the candidate summary generator aggregates EDU representation vectors to generate several candidate summaries with varying lengths by specifying different $k$ values. Different from the previous top-$k$ strategy where $k$ is a fixed value, multiple $k$ values are provided to the proposed model, allowing different numbers of EDUs being extracted to form different candidate summaries with varying lengths for the same document. Lastly, the document-level block encodes each candidate summary and selects one of the candidate summaries as the final model output. In this way, the proposed model decides the most suitable summary length, i.e., $k$, for each document.


\paragraph{EDU-level Block} Given input document \(\mathcal{D}=[w_{11},\dots,w_{m{n_m}}]\) where \(w_{ij}\) denotes \(j\)-th word in \(i\)-th EDU, [CLS] and [SEP] tokens are inserted into \(\mathcal{D}\) at the start and end of each EDU. We adapt the pre-trained Transformer-based language model (PLM) as the EDU encoder, e.g., RoBERTa. The hidden states of [CLS] tokens derived from the PLM are taken as EDU representations, i.e., $\mathbf{edu^E}$ in Equation (\ref{equation: plm}). A classification layer is further applied on EDU representations to predict probabilities, i.e., $\mathbf{P}$ in Equation (\ref{equation: classifier}).
\begin{gather}
    [edu_1^E, \dots, edu_m^E]=\mathbf{PLM}_{\theta}(\mathcal{D}) \label{equation: plm} \\
     P_i(y_i=1)=\sigma(\mathbf{W}^c edu_i^E + \mathbf{b}^c), \label{equation: classifier}
\end{gather}
where $\theta$ is the set of all trainable parameters in PLM; \(\mathbf{W}^c\) and \(\mathbf{b}^c\) are trainable parameters in classification layer, and \(\sigma(\cdot)\) denotes sigmoid function.

\paragraph{Candidate Summary Generator} Given a pre-defined extraction lengths set \(\mathcal{K}=[k_1, \dots, k_c]\), the $s$-th candidate summary, $cand_s$, consists of EDUs whose probabilities are in top-${k_s}(\mathbf{P})$, i.e., $[edu_{i_1},\dots,edu_{i_j},\dots,edu_{i_k}]$ where \(i_j \leq m\) and \(P_{i_j} \in\) top-\({k_s}(\mathbf{P}), j=1,2,...,k_s\). The initial representation vector, $cand_s^C$, for $cand_s$ is the concatenation of representation vectors of EDUs in it. The initial document representation vector, $\mathcal{D}^C$, is aggregated from the representation vectors of all EDUs.


\paragraph{Document-level Block} Multiple Transformer encoder layers (MTL) are stacked to encode document-level information for document $\mathcal{D}^C$, and all candidate summaries, e.g., $cand_s^C$, separately, and generate $\mathcal{D}^D$ and $cand_s^D$ in Equation (\ref{equation: MLT}). Then cosine similarity, i.e., $sim_s$ in Equation (\ref{equation: cosine}), is computed between the encoded document representation and the encoded $s$-th candidate summary representation. The candidate summary with the highest similarity with the document is taken as the final model-generated summary.
\begin{gather}
    \resizebox{.85\hsize}{!}{$[\mathcal{D}^D, cand^D_s]=[\mathbf{MTL}_\eta(\mathcal{D}^C), \mathbf{MTL}_\eta(cand^C_s)]$} \label{equation: MLT} \\
    sim_s=\mathbf{cosine}(\mathcal{D}^D,cand^D_s) \label{equation: cosine},
\end{gather}
where $\eta$ is the set of trainable parameters in MTL.

\paragraph{Training} Algorithm \ref{algo:model learning} summarizes the model learning procedure. The model encodes EDUs in the document and predicts the probability for each EDU (lines 1-2), generates indices of EDUs for candidate summaries with different lengths which are derived from different $k$ values (lines 3-4), encodes the whole document and candidate summaries and calculates similarity scores (lines 7-10), and selects the best candidate summary (line 16) in an end-to-end manner. Inspired by \citet{zhong-etal-2020-extractive} that the candidate summary having a higher ROUGE score with the reference summary is expected to have a higher similarity score with the whole document, during training, ROUGE scores for each $(\mathcal{R},cand_s)$ pair are calculated and used to sort the set $\mathcal{C}$ in descending order (lines 5-6) to align with the loss function in Equation (\ref{equation: con1}). Besides, to better emphasize those important EDUs, the EDU-level oracle summary, denoted as $cand_{gt}$ here, is introduced to the training process and assumed to have the highest ROUGE score (lines 12-13).

\begin{algorithm}[t]
\caption{Model Learning Algorithm}
\label{algo:model learning}
\begin{algorithmic}[1]
\Require{$\mathcal{D\vert}_1^m, \mathcal{K}\vert_1^c, L\vert_1^m$}
\Ensure{candSumIdx}
\State eduRep$\vert_1^m \gets$ \textit{PLM}$_\theta$($\mathcal{D}$) 
\State $P\vert_1^m \gets $ \textit{classification}$_{w,b}$(eduRep$\vert_1^m$)
\For{$i \gets 1$ to $c$}
  \State selIdx$_i\gets$ indices of top-$\mathcal{K}_i$($P\vert_1^m$)
\EndFor
\If{training}
  \State selIdx$\vert_1^c\gets$ sort based on ROUGE scores
\EndIf
\State docRep $\gets$ \textit{MTL}$_\eta$(eduRep$\vert_1^m$)
\For{$j \gets 1$ to $c$}
  \State candRep$_j \gets$ \textit{MTL}$_\eta$(eduRep\textsubscript{$\in$selIdx$_j$})
  \State sim$_j \gets$ \textit{cosine}(docRep, candRep$_j$)
\EndFor
\If{training}
  \State gtIdx $\gets$ indices of 1 in $L\vert_1^m$
  \State sim$_{gt} \gets$ repeat 9-10
  \State $\mathcal{L} \gets$ loss from $P\vert_1^m, L\vert_1^m, sim\vert_1^c, sim_{gt}$
  \State $\theta,w,b,\eta \gets$ parameters updated by $\mathcal{L}$
\EndIf
\State candSumIdx $\gets$ selIdx\textsubscript{index\_\textit{max}(sim$\vert_1^c$)}
\State \textbf{return} candSumIdx
\end{algorithmic}
\end{algorithm}

\subsection{Objective Function}
Binary cross entropy is calculated on the outputs of the classification layer in the EDU-level block, as in Equation (\ref{equation: bce}). Contrastive learning loss is calculated on the outputs of the similarity layer in the document-level block, as in Equations (\ref{equation: con1}-\ref{equation: con2}). The final training loss $\mathcal{L}$ in Equation (\ref{equation: loss}) is calculated as a weighted summation between them. 
\begin{equation}
    \mathcal{L}=\mathcal{L}_{bce} + \rho * \mathcal{L}_{con}, \label{equation: loss}
\end{equation}
where
\begin{gather}
    \resizebox{.85\hsize}{!}{$\mathcal{L}_{bce}=-\sum_{i=1}^{m}(l_i log(P_i) + (1-l_i) log(1-P_i))$} \label{equation: bce} \\
    \mathcal{L}_{con}=\mathcal{L}_1 + \mathcal{L}_2 \label{equation: con1},
\end{gather}
where 
\begin{gather}
    \mathcal{L}_{1}=\sum_{s=1}^c max(0, sim_s-sim_{gt}+\gamma_1) \\
    \resizebox{.85\hsize}{!}{$\mathcal{L}_{2}=\sum_{i<j}^cmax(0, sim_j-sim_i +(j-i) * \gamma_2)$} \label{equation: con2}
\end{gather}

\section{Experiments}

\subsection{Datasets}
\textbf{CNN/DailyMail} \citep{Hermann2015TeachingComprehend} is the most commonly used news dataset for the extractive task with human-written highlights as reference summary. The non-anonymized version was used in our experiments. \textbf{XSum} \citep{narayan-etal-2018-dont} is another news dataset with the first introductory sentence in the article as the reference summary. \textbf{Reddit} \citep{Kim2019AbstractiveNetworks} is a dataset crawled from the social media forum with the content in the section TL;DR as the reference summary. Experiments were conducted on the TIFU-long version. \textbf{WikiHow} \citep{Koupaee2018WikiHow:Dataset} is a dataset crawled from the question-answering website with the first sentence in each paragraph as the reference summary. \textbf{Multi-News} \citep{fabbri-etal-2019-multi} is a multi-document dataset with one summary for a cluster of documents. We follow \citeposs{zhong-etal-2020-extractive} setting to split Reddit and Multi-News datasets and concatenate multiple documents into one single document. The detailed statistics of the five datasets in our experiments can be found in Appendix \ref{appendix:datatset stats}.

\subsection{Baselines}
Various extractive models are selected as baselines. \textbf{HETFORMER} \citep{liu-etal-2021-hetformer} modifies Longformer with longer input lengths to implement multi-granularity attention and selects sentences. Among models generating summaries with varying lengths, \textbf{MATCHSUM}  \citep{zhong-etal-2020-extractive} selects among a set of candidate summaries derived from a trained sentence-level extractive model; \textbf{HAHSUM} \citep{jia-etal-2020-neural} transforms a document into a heterogeneous hierarchical graph and flexibly selects sentences based on a threshold. Among models with sub-sentential segments as input, the \textbf{Proposed} model by \citet{huang-kurohashi-2021-extractive} is another Longformer-based model but extracts EDUs based on the constructed heterogeneous graph; \textbf{DISCOBERT} \citep{xu-etal-2020-discourse} and \textbf{D-SUM} \citep{liu-chen-2019-exploiting} are models extracting discourse-level textual segments but they differ in whether integrating GNN into the model. \textbf{SGSUM} \citep{chen-etal-2021-sgsum} is a multi-document model by encoding all documents within one cluster individually and selecting the best sub-graph. \textbf{FAR} \citep{liang-etal-2021-improving} is an unsupervised ranking model considering facet-specific information.

\subsection{Experimental Setting}

EDU segmentation of sentences in the document is conducted by NeuralEDUSegmentation\footnote{\url {https://github.com/PKU-TANGENT/NeuralEDUSeg}} \citep{wang-etal-2018-toward}. To facilitate the training process, the calculation of ROUGE scores is avoided by pre-selecting the set of candidate summaries based on the predicted probabilities by the fine-tuned RoBERTa on the extractive task for each dataset. The pre-trained ``roberta-base'' or ``bart-base'' is adapted as the EDU encoder and enlarged to handle the first 768 BPEs of each document. The number of Transformer encoder layers is 4 by default. Following \citet{liu-lapata-2019-text}, a similar greedy algorithm is applied to generate ground truth labels for EDUs (also for oracle summaries in Section \ref{section:empirial justification}) and the pseudo-code is in Appendix \ref{appendix: greedy algo}. The trigram strategy is applied when forming the final EDU-constituent summary during validating and testing.

We follow \citeposs{zhong-etal-2020-extractive} setting to set up $\gamma_1 = 0$ and $\gamma_2 = 0.01$. $\rho$ is set as 100 based on our observation during training. Adam optimizer is used. The batch size is 5 to fit the GPU memory limit during training and 60 during validating or testing. Every 6k steps are defined as one epoch; the training process could take up to 100 epochs and early stopping is activated with patience as 10 epochs and R-2 as the metric. Experiments are conducted on a single Nvidia-v100-16GB GPU. The F1-scores of ROUGE-1/2/L\footnote{\url {https://github.com/bheinzerling/pyrouge}} \citep{lin-2004-rouge} are taken as the automatic evaluation metrics. More details are provided in Appendix \ref{appendix: more settings}.


\subsection{Experimental Results}

\begin{table}[t]
    \centering
    \resizebox{\columnwidth}{!}
    {%
    \begin{tabular}{lccc}
        \hline
        Model & R-1 & R-2 & R-L \\
        \hline
        ORACLE (EDU) & 62.50 & 38.67 & 60.16 \\
        ORACLE (sentence) & 55.31 & 32.73 & 51.63 \\
        LEAD-3 (sentence) & 39.96 & 17.39 & 36.27 \\
        \hline
        \textsc{D-Sum} \citep{liu-chen-2019-exploiting} & 42.78 & 20.23 & - \\
        \textsc{DiscoBERT} \citep{xu-etal-2020-discourse} & 43.77 & 20.85 & 40.67 \\
        Proposed \citep{huang-kurohashi-2021-extractive} & 43.61 & 20.81 & 41.12 \\
        \textsc{HAHSum} \citep{jia-etal-2020-neural} & 44.68 & 21.30 & 40.75 \\
        \textsc{MatchSum} \citep{zhong-etal-2020-extractive} & 44.41 & 20.86 & 40.55 \\
        \textsc{HetFormer} \citep{liu-etal-2021-hetformer} & 44.55 & 20.82 & 40.37 \\
        \textsc{FAR} \citep{liang-etal-2021-improving} & 40.83 & 17.85 & 36.91 \\
        \hline
        $\textsc{EDU-VL}_\textsc{RoBERTa}$ & \textbf{44.80} & \textbf{21.66} & \textbf{42.56} \\
        $\textsc{EUD-VL}_\textsc{BART}$ & 44.70 & 21.63 & 42.46 \\
        \hline
    \end{tabular} %
    }
    \caption{F1-scores on CNN/DailyMail test dataset.}
    \label{table: cnndm results}
\end{table}

\paragraph{CNN/DailyMail} Table \ref{table: cnndm results} shows the results. The top section includes F1-scores of oracle summaries and the Lead-3 method. The second section presents the F1-scores reported in the original papers of all baselines. The last section lists the F1-scores of our proposed model. 

Our proposed model outperforms the unsupervised baseline, \textsc{FAR}, by a large margin, aligning with the observation from other supervised baselines. Compared with discourse-level baselines, i.e., \textsc{D-Sum}, \textsc{DiscoBERT} and Proposed, our proposed model achieves an improvement of at least 1.03/0.81/1.44 on R-1/2/L. When compared against other two varying lengths-enabled models, i.e., \textsc{HAHSum} and \textsc{MatchSum}, our proposed model achieves better R-1 result on a small scale (0.12) and R-2/L on a larger scale (0.8/1.81). Our proposed model also beats \textsc{HetFormer} which allows longer input length by a similar scale pattern. It is observed that the RoBERTa version of our proposed model performs slightly better than the BART version. The experimental results suggest that our proposed model achieves better performance than all baselines on the R-1/2/L.

\begin{table}[t]
    \centering
    \resizebox{\columnwidth}{!}
    {%
    \begin{tabular}{lccc}
        \hline
        Model & R-1 & R-2 & R-L \\
        \hline
        \multicolumn{4}{c}{\textbf{XSum}} \\
        \hline
        ORACLE (EDU) & 36.16 & 11.74 & 31.02 \\
        ORACLE (sentence) & 29.11 & 8.66 & 22.29 \\
        LEAD-3 (sentence) & 19.41 & 2.65 & 15.05 \\
        \hline
        \textsc{MatchSum} \citep{zhong-etal-2020-extractive} & 24.86 & 4.66 & 18.41 \\
        \hline
        $\textsc{EDU-VL}_\textsc{RoBERTa}$ & \textbf{26.48} & 5.74 & 22.33 \\
        $\textsc{EDU-VL}_\textsc{BART}$ & 26.43 & \textbf{5.78} & \textbf{22.35} \\
        \hline
        \multicolumn{4}{c}{\textbf{Reddit}} \\
        \hline
        ORACLE (EDU) & 44.49 & 18.53 & 38.87 \\
        ORACLE (sentence) & 34.36 & 12.97 & 26.98 \\
        LEAD-3 (sentence) & 18.39 & 3.01 & 14.12 \\
        \hline
        \textsc{MatchSum} \citep{zhong-etal-2020-extractive} & 25.09 & 6.17 & 20.13 \\
        \hline
        $\textsc{EUD-VL}_\textsc{RoBERTa}$ & \textbf{27.04} & 6.87 & 22.64 \\
        $\textsc{EDU-VL}_\textsc{BART}$ & 27.01 & \textbf{7.06} & \textbf{22.70} \\
        \hline
    \end{tabular} %
    }
    \caption{F1-score results on test dataset of XSum and Reddit. The number of Transformer encoder layers in BART version of XSum is 6 and 2 for both versions of Reddit.}
    \label{table: xsum and reddit results}
\end{table}

\paragraph{XSum and Reddit} The results in Table \ref{table: xsum and reddit results} show that our proposed model outperforms the baseline model, \textsc{MatchSum}, by a large margin on all three metrics (1.57/1.12/3.94 and 1.92/0.89/2.57 on R-1/2/L for XSum and Reddit, respectively). The RoBERTa version of our model only achieves slightly better result on R-1 than the BART version.

\begin{table}[t]
    \centering
    \resizebox{\columnwidth}{!}
    {%
    \begin{tabular}{lccc}
        \hline
        Model & R-1 & R-2 & R-L \\
        \hline
        \multicolumn{4}{c}{\textbf{WikiHow}} \\
        \hline
        ORACLE (EDU) & 44.13 & 17.90 & 42.38 \\
        ORACLE (sentence) & 37.89 & 13.80 & 35.13 \\
        LEAD-3 (sentence) & 23.97 & 5.37 & 22.22 \\
        \hline
        \textsc{FAR} \citep{liang-etal-2021-improving} & 27.54 & 6.17 & 25.46 \\
        \textsc{MatchSum} \citep{zhong-etal-2020-extractive} & 31.85 & 8.98 & 29.58 \\
        \hline
        $\textsc{EDU-VL}_\textsc{RoBERTa}$ & 33.94 & 10.31 & 32.55 \\
        $\textsc{EDU-VL}_\textsc{BART}$ & \textbf{34.01} & \textbf{10.45} & \textbf{32.66} \\
        \hline
        \multicolumn{4}{c}{\textbf{Multi-News}} \\
        \hline
        ORACLE (EDU) & 51.60 & 24.24 & 48.92 \\
        ORACLE (sentence) & 49.87 & 22.43 & 45.18 \\
        LEAD-3 (sentence) & 28.40 & 8.63 & 24.93 \\
        \hline
        \textsc{HetFormer} \citep{liu-etal-2021-hetformer} & 46.21 & 17.49 & 42.43 \\
        \textsc{SgSum} \citep{chen-etal-2021-sgsum} & 47.53 & \textbf{18.75} & 43.31 \\
        \textsc{FAR} \citep{liang-etal-2021-improving} & 43.48 & 16.87 & 44.00 \\
        \textsc{MatchSum} \citep{zhong-etal-2020-extractive} & 46.20 & 16.51 & 41.89 \\
        \hline
        $\textsc{EDU-VL}_\textsc{RoBERTa}$ & 46.82 & 17.05 & 44.36 \\
        $\textsc{EDU-VL}_\textsc{BART}$ & \textbf{47.56} & 17.64 & \textbf{45.05} \\
        \hline
    \end{tabular} %
    }
    \caption{F1-score results on test dataset of WikiHow and Multi-News.}
    \label{table: wikihow and multinews results}
\end{table}

\paragraph{WikiHow and Multi-News} As shown in Table \ref{table: wikihow and multinews results}, our proposed model achieves significantly better performance on WikiHow dataset, beating both \textsc{MatchSum} and \textsc{FAR} by at least 2.16/1.47/3.08 on R-1/2/L. For the Multi-News dataset, our proposed model outperforms \textsc{HetFormer}, \textsc{MatchSum} and \textsc{FAR}. It is noteworthy that \textsc{SgSum} is initially designed to incorporate multiple documents, meaning that its input document is more complete than ours. Though our proposed model underperforms \textsc{SgSum} on R-2, our proposed model achieves comparable result on R-1 and better result on R-L. The BART version of our proposed model outperforms the RoBERTa version on all three metrics on both datasets. To sum up, our proposed model performs better on WikiHow dataset and comparably on Multi-News dataset when compared against the corresponding state-of-the-art baselines.

\subsection{Analysis}

\paragraph{Ablation Analysis}

\begin{table}[t]
    \centering
    \begin{tabular}{lccc}
        \hline
        Model & R-1 & R-2 & R-L \\
        \hline
        $\textsc{EDU-VL}_\textsc{RoBERTa}$ & \textbf{44.80} & \textbf{21.66} & \textbf{42.56} \\
        w/o EDU & 43.89 & 20.79 & 40.18 \\
        w/o VL & 44.32 & 21.38 & 42.12 \\
        \hline
    \end{tabular}
    \caption{Ablation analysis on test dataset of CNN/DM.}
    \label{table: cnndm ablation results}
\end{table}

We further conduct ablation analysis by removing specific characteristics in our model and the result is presented in Table \ref{table: cnndm ablation results}. Both letting the model extract sentences under the same architecture and removing the document-block to disable the varying lengths characteristic reduce model performance on all three metrics. A larger decrease is observed in the sentence-level model.

\paragraph{Human Evaluation} We randomly sample 50 summaries generated by our model from the CNN/DailyMail test dataset and conduct detailed qualitative analysis. For each summary, we combine EDUs from the same sentence together as one textual segment. Then referring to the dependency tree of the corresponding sentence, we evaluate the syntactical completeness of the extracted textual segment. Out of 221 extracted textual segments in all 50 summaries, 68\% are syntactically complete and 32\% are not. It is noteworthy that about half of those incomplete ones are subordinate clauses, whose syntax structure is close to being complete. Out of these complete ones, 44.7\% are the whole sentence itself because all EDUs in that sentence are extracted; 55.3\% maintain complete syntax structure after dropping some EDU(s) in that sentence (as the example shown in Table \ref{tab:model example}). Therefore, it is safe to believe that even sentences split into multiple EDUs, the model is capable to maintain the syntax structure by choosing multiple EDUs in a sentence and in some cases, filtering out some redundant information without breaking the completeness of the syntax.



\paragraph{Generated Summary Examples}
Table \ref{tab:model example} provides an example of a summary generated by our proposed model, which illustrates that the model manages to selectively drop redundant information in sentences by operating on the EDU-level while maintaining an informative and readable summary.

\begin{table}[t]
    \centering
    \begin{tabular}{p{0.9\linewidth}}
    \hline
        \textit{\textbf{Document:}} (...) [\textcolor{blue}{\textit{Arnold Breitenbach of St. George wanted to get `CIB-69' put on a license plate,}}]\textsubscript{21} [the Spectrum newspaper of St. George reported.]\textsubscript{22} [\textcolor{brown}{\textit{That would have commemorated both Breitenbach getting the Purple Heart in 1969 and his Combat Infantryman's Badge,}}]\textsubscript{31} [according to the newspaper.]\textsubscript{32} (...) [\textcolor{orange}{\textit{The Utah DMV denied his request,}}]\textsubscript{51} [\textcolor{orange}{\textit{citing state regulations}}]\textsubscript{52} [\textcolor{orange}{\textit{prohibiting the use of the number 69}}]\textsubscript{53} [\textcolor{orange}{\textit{because of its sexual connotations}}]\textsubscript{54} (...) \\
        \hline
        \textit{\textbf{Reference Summary: }} \textcolor{blue}{Arnold Breitenbach of St. George, Utah, wanted to get `CIB-69' put on a license plate.} \textcolor{brown}{That would have commemorated both Breitenbach getting the Purple Heart in 1969 and his Combat Infantryman's Badge.} \textcolor{orange}{The Utah DMV denied his request, citing state regulations prohibiting the use of the number 69 because of its sexual connotations.}
 \\
        \hline
    \end{tabular}
    \caption{Example from model-generated summary. Content within [] represents an EDU and subscript number $ij$ indicates it is the $j$-th EDU in the $i$-th sentence in the document. Each color represents information in a sentence in reference summary. Italic denotes content selected by \textit{our proposed model}.}
    \label{tab:model example}
\end{table}

\section{Conclusion}
In this paper, we verify and quantify the argument that the EDU-level summary achieves higher automatic evaluation scores than sentence-level summary from both theoretical and experimental perspectives. We further propose an EDU-level extractive summarization model and develop its learning algorithm, which generates summaries with different lengths for different documents. The experimental results demonstrate that our model achieves superior performance on four single-document summarization datasets and comparable performance for multi-document summarization with direct comparison with the multi-document model. In the future, we will explore integrating the EDU-level summary generated by our model into the abstractive summarization model.

\section*{Limitations}
Though EDU is defined as a clause in a sentence, current EDU segmenters are still underdeveloped due to the limited training dataset and usually split a sentence into consecutive EDUs, which breaks the syntactic structure. Occasionally some extracted EDUs from a sentence fail to recover a complete syntactic structure. Therefore, a more sophisticated segmenter could further improve the segmentation, or some post-processing treatments could be developed to address such a potential issue specifically.



\section*{Acknowledgements}
We would like to acknowledge the assistance given by Research IT and the use of the Computational Shared Facility at The University of Manchester. We thank the anonymous reviewers for their helpful comments.

\bibliography{anthology, references}

\begin{thebibliography}{39}
\expandafter\ifx\csname natexlab\endcsname\relax\def\natexlab#1{#1}\fi

\bibitem[{Alonso~i Alemany and
  Fuentes~Fort(2003)}]{AlonsoiAlemany2003IntegratingSummarization}
Laura Alonso~i Alemany and Maria Fuentes~Fort. 2003.
\newblock \href {https://doi.org/10.3115/1067737.1067739} {{Integrating
  cohesion and coherence for automatic summarization}}.
\newblock In \emph{Proceedings of EACL2003}, page~1.

\bibitem[{Chen et~al.(2021)Chen, Li, Liu, Xiao, Wu, and
  Wang}]{chen-etal-2021-sgsum}
Moye Chen, Wei Li, Jiachen Liu, Xinyan Xiao, Hua Wu, and Haifeng Wang. 2021.
\newblock \href {https://doi.org/10.18653/v1/2021.emnlp-main.333}
  {{S}g{S}um:transforming multi-document summarization into sub-graph
  selection}.
\newblock In \emph{Proceedings of the 2021 Conference on Empirical Methods in
  Natural Language Processing}, pages 4063--4074, Online and Punta Cana,
  Dominican Republic. Association for Computational Linguistics.

\bibitem[{Cho et~al.(2020)Cho, Song, Li, Yu, Foroosh, and
  Liu}]{cho-etal-2020-better}
Sangwoo Cho, Kaiqiang Song, Chen Li, Dong Yu, Hassan Foroosh, and Fei Liu.
  2020.
\newblock \href {https://doi.org/10.18653/v1/2020.emnlp-main.509} {Better
  highlighting: Creating sub-sentence summary highlights}.
\newblock In \emph{Proceedings of the 2020 Conference on Empirical Methods in
  Natural Language Processing (EMNLP)}, pages 6282--6300, Online. Association
  for Computational Linguistics.

\bibitem[{Cui et~al.(2020)Cui, Hu, and Liu}]{cui-etal-2020-enhancing}
Peng Cui, Le~Hu, and Yuanchao Liu. 2020.
\newblock \href {https://doi.org/10.18653/v1/2020.coling-main.468} {Enhancing
  extractive text summarization with topic-aware graph neural networks}.
\newblock In \emph{Proceedings of the 28th International Conference on
  Computational Linguistics}, pages 5360--5371, Barcelona, Spain (Online).
  International Committee on Computational Linguistics.

\bibitem[{Devlin et~al.(2019)Devlin, Chang, Lee, and
  Toutanova}]{devlin-etal-2019-bert}
Jacob Devlin, Ming-Wei Chang, Kenton Lee, and Kristina Toutanova. 2019.
\newblock \href {https://doi.org/10.18653/v1/N19-1423} {{BERT}: Pre-training of
  deep bidirectional transformers for language understanding}.
\newblock In \emph{Proceedings of the 2019 Conference of the North {A}merican
  Chapter of the Association for Computational Linguistics: Human Language
  Technologies, Volume 1 (Long and Short Papers)}, pages 4171--4186,
  Minneapolis, Minnesota. Association for Computational Linguistics.

\bibitem[{Dong et~al.(2018)Dong, Shen, Crawford, van Hoof, and
  Cheung}]{dong-etal-2018-banditsum}
Yue Dong, Yikang Shen, Eric Crawford, Herke van Hoof, and Jackie Chi~Kit
  Cheung. 2018.
\newblock \href {https://doi.org/10.18653/v1/D18-1409} {{B}andit{S}um:
  Extractive summarization as a contextual bandit}.
\newblock In \emph{Proceedings of the 2018 Conference on Empirical Methods in
  Natural Language Processing}, pages 3739--3748, Brussels, Belgium.
  Association for Computational Linguistics.

\bibitem[{Ernst et~al.(2022)Ernst, Caciularu, Shapira, Pasunuru, Bansal,
  Goldberger, and Dagan}]{Ernst2022Proposition-LevelSummarization}
Ori Ernst, Avi Caciularu, Ori Shapira, Ramakanth Pasunuru, Mohit Bansal, Jacob
  Goldberger, and Ido Dagan. 2022.
\newblock \href {https://doi.org/10.18653/v1/2022.naacl-main.128}
  {{Proposition-Level Clustering for Multi-Document Summarization}}.
\newblock In \emph{Proceedings of the 2022 Conference of the North American
  Chapter of the Association for Computational Linguistics: Human Language
  Technologies}, pages 1765--1779, Stroudsburg, PA, USA. Association for
  Computational Linguistics.

\bibitem[{Fabbri et~al.(2019)Fabbri, Li, She, Li, and
  Radev}]{fabbri-etal-2019-multi}
Alexander Fabbri, Irene Li, Tianwei She, Suyi Li, and Dragomir Radev. 2019.
\newblock \href {https://doi.org/10.18653/v1/P19-1102} {Multi-news: A
  large-scale multi-document summarization dataset and abstractive hierarchical
  model}.
\newblock In \emph{Proceedings of the 57th Annual Meeting of the Association
  for Computational Linguistics}, pages 1074--1084, Florence, Italy.
  Association for Computational Linguistics.

\bibitem[{Gu et~al.(2022)Gu, Ash, and Hahnloser}]{gu-etal-2022-memsum}
Nianlong Gu, Elliott Ash, and Richard Hahnloser. 2022.
\newblock \href {https://doi.org/10.18653/v1/2022.acl-long.450} {{M}em{S}um:
  Extractive summarization of long documents using multi-step episodic {M}arkov
  decision processes}.
\newblock In \emph{Proceedings of the 60th Annual Meeting of the Association
  for Computational Linguistics (Volume 1: Long Papers)}, pages 6507--6522,
  Dublin, Ireland. Association for Computational Linguistics.

\bibitem[{Hermann et~al.(2015)Hermann, Ko{\v{c}}isk{\'{y}}, Grefenstette,
  Espeholt, Kay, Suleyman, and Blunsom}]{Hermann2015TeachingComprehend}
Karl~Moritz Hermann, Tomáš Ko{\v{c}}isk{\'{y}}, Edward Grefenstette, Lasse
  Espeholt, Will Kay, Mustafa Suleyman, and Phil Blunsom. 2015.
\newblock {Teaching machines to read and comprehend}.
\newblock \emph{Advances in Neural Information Processing Systems},
  2015-Janua:1693--1701.

\bibitem[{Huang and Kurohashi(2021)}]{huang-kurohashi-2021-extractive}
Yin~Jou Huang and Sadao Kurohashi. 2021.
\newblock \href {https://doi.org/10.18653/v1/2021.eacl-main.265} {Extractive
  summarization considering discourse and coreference relations based on
  heterogeneous graph}.
\newblock In \emph{Proceedings of the 16th Conference of the European Chapter
  of the Association for Computational Linguistics: Main Volume}, pages
  3046--3052, Online. Association for Computational Linguistics.

\bibitem[{Jia et~al.(2020)Jia, Cao, Tang, Fang, Cao, and
  Wang}]{jia-etal-2020-neural}
Ruipeng Jia, Yanan Cao, Hengzhu Tang, Fang Fang, Cong Cao, and Shi Wang. 2020.
\newblock \href {https://doi.org/10.18653/v1/2020.emnlp-main.295} {Neural
  extractive summarization with hierarchical attentive heterogeneous graph
  network}.
\newblock In \emph{Proceedings of the 2020 Conference on Empirical Methods in
  Natural Language Processing (EMNLP)}, pages 3622--3631, Online. Association
  for Computational Linguistics.

\bibitem[{Jing et~al.(2021)Jing, You, Yang, Fan, and
  Tong}]{jing-etal-2021-multiplex}
Baoyu Jing, Zeyu You, Tao Yang, Wei Fan, and Hanghang Tong. 2021.
\newblock \href {https://doi.org/10.18653/v1/2021.emnlp-main.11} {Multiplex
  graph neural network for extractive text summarization}.
\newblock In \emph{Proceedings of the 2021 Conference on Empirical Methods in
  Natural Language Processing}, pages 133--139, Online and Punta Cana,
  Dominican Republic. Association for Computational Linguistics.

\bibitem[{Kim et~al.(2019)Kim, Kim, and Kim}]{Kim2019AbstractiveNetworks}
Byeongchang Kim, Hyunwoo Kim, and Gunhee Kim. 2019.
\newblock {Abstractive summarization of reddit posts with multi-level memory
  networks}.
\newblock In \emph{NAACL HLT 2019 - 2019 Conference of the North American
  Chapter of the Association for Computational Linguistics: Human Language
  Technologies - Proceedings of the Conference}, volume~1, pages 2519--2531.

\bibitem[{Koupaee and Wang(2018)}]{Koupaee2018WikiHow:Dataset}
Mahnaz Koupaee and William~Yang Wang. 2018.
\newblock \href {http://arxiv.org/abs/1810.09305} {{WikiHow: A Large Scale Text
  Summarization Dataset}}.
\newblock In \emph{arXiv preprint arXiv:1810.09305}, pages 1--5.

\bibitem[{Kwon et~al.(2021)Kwon, Kobayashi, Kamigaito, and
  Okumura}]{kwon-etal-2021-considering}
Jingun Kwon, Naoki Kobayashi, Hidetaka Kamigaito, and Manabu Okumura. 2021.
\newblock \href {https://doi.org/10.18653/v1/2021.emnlp-main.330} {Considering
  nested tree structure in sentence extractive summarization with pre-trained
  transformer}.
\newblock In \emph{Proceedings of the 2021 Conference on Empirical Methods in
  Natural Language Processing}, pages 4039--4044, Online and Punta Cana,
  Dominican Republic. Association for Computational Linguistics.

\bibitem[{Lewis et~al.(2020)Lewis, Liu, Goyal, Ghazvininejad, Mohamed, Levy,
  Stoyanov, and Zettlemoyer}]{lewis-etal-2020-bart}
Mike Lewis, Yinhan Liu, Naman Goyal, Marjan Ghazvininejad, Abdelrahman Mohamed,
  Omer Levy, Veselin Stoyanov, and Luke Zettlemoyer. 2020.
\newblock \href {https://doi.org/10.18653/v1/2020.acl-main.703} {{BART}:
  Denoising sequence-to-sequence pre-training for natural language generation,
  translation, and comprehension}.
\newblock In \emph{Proceedings of the 58th Annual Meeting of the Association
  for Computational Linguistics}, pages 7871--7880, Online. Association for
  Computational Linguistics.

\bibitem[{Li et~al.(2016)Li, Thadani, and Stent}]{li-etal-2016-role}
Junyi~Jessy Li, Kapil Thadani, and Amanda Stent. 2016.
\newblock \href {https://doi.org/10.18653/v1/W16-3617} {The role of discourse
  units in near-extractive summarization}.
\newblock In \emph{Proceedings of the 17th Annual Meeting of the Special
  Interest Group on Discourse and Dialogue}, pages 137--147, Los Angeles.
  Association for Computational Linguistics.

\bibitem[{Liang et~al.(2021)Liang, Wu, Li, and Li}]{liang-etal-2021-improving}
Xinnian Liang, Shuangzhi Wu, Mu~Li, and Zhoujun Li. 2021.
\newblock \href {https://doi.org/10.18653/v1/2021.findings-acl.147} {Improving
  unsupervised extractive summarization with facet-aware modeling}.
\newblock In \emph{Findings of the Association for Computational Linguistics:
  ACL-IJCNLP 2021}, pages 1685--1697, Online. Association for Computational
  Linguistics.

\bibitem[{Lin(2004)}]{lin-2004-rouge}
Chin-Yew Lin. 2004.
\newblock \href {https://aclanthology.org/W04-1013} {{ROUGE}: A package for
  automatic evaluation of summaries}.
\newblock In \emph{Text Summarization Branches Out}, pages 74--81, Barcelona,
  Spain. Association for Computational Linguistics.

\bibitem[{Liu and Lapata(2019)}]{liu-lapata-2019-text}
Yang Liu and Mirella Lapata. 2019.
\newblock \href {https://doi.org/10.18653/v1/D19-1387} {Text summarization with
  pretrained encoders}.
\newblock In \emph{Proceedings of the 2019 Conference on Empirical Methods in
  Natural Language Processing and the 9th International Joint Conference on
  Natural Language Processing (EMNLP-IJCNLP)}, pages 3730--3740, Hong Kong,
  China. Association for Computational Linguistics.

\bibitem[{Liu et~al.(2021)Liu, Zhang, Wan, Xia, He, and
  Yu}]{liu-etal-2021-hetformer}
Ye~Liu, Jianguo Zhang, Yao Wan, Congying Xia, Lifang He, and Philip Yu. 2021.
\newblock \href {https://doi.org/10.18653/v1/2021.emnlp-main.13} {{HETFORMER}:
  Heterogeneous transformer with sparse attention for long-text extractive
  summarization}.
\newblock In \emph{Proceedings of the 2021 Conference on Empirical Methods in
  Natural Language Processing}, pages 146--154, Online and Punta Cana,
  Dominican Republic. Association for Computational Linguistics.

\bibitem[{Liu et~al.(2019)Liu, Ott, Goyal, Du, Joshi, Chen, Levy, Lewis,
  Zettlemoyer, and Stoyanov}]{Liu2019RoBERTa:Approach}
Yinhan Liu, Myle Ott, Naman Goyal, Jingfei Du, Mandar Joshi, Danqi Chen, Omer
  Levy, Mike Lewis, Luke Zettlemoyer, and Veselin Stoyanov. 2019.
\newblock \href {http://arxiv.org/abs/1907.11692} {{RoBERTa: A Robustly
  Optimized BERT Pretraining Approach}}.
\newblock \emph{arXiv preprint arXiv:1907.11692}.

\bibitem[{Liu and Chen(2019)}]{liu-chen-2019-exploiting}
Zhengyuan Liu and Nancy Chen. 2019.
\newblock \href {https://doi.org/10.18653/v1/D19-5415} {Exploiting
  discourse-level segmentation for extractive summarization}.
\newblock In \emph{Proceedings of the 2nd Workshop on New Frontiers in
  Summarization}, pages 116--121, Hong Kong, China. Association for
  Computational Linguistics.

\bibitem[{Lunh(1958)}]{Lunh1958TheAbstracts}
H.~P. Lunh. 1958.
\newblock \href
  {http://www.di.ubi.pt/~jpaulo/competence/general/(1958)Luhn.pdf} {{The
  Automatic Creation of Literature Abstracts}}.
\newblock \emph{IBM Journal of Research Development}, 2(2):159--165.

\bibitem[{Mann and Thompson(1988)}]{Mann1988RhetoricalOrganization}
William~C. Mann and Sandra~A. Thompson. 1988.
\newblock \href {https://doi.org/10.1515/text.1.1988.8.3.243} {{Rhetorical
  Structure Theory: Toward a functional theory of text organization}}.
\newblock \emph{Text-interdisciplinary Journal for the Study of Discourse},
  8(3):243--281.

\bibitem[{Marcu(1999)}]{Marcu1999DiscourseText}
Daniel Marcu. 1999.
\newblock {Discourse trees are good indicators of importance in text}.
\newblock In \emph{Advances in automatic text summarization}, pages 123--136.

\bibitem[{Nallapati et~al.(2017)Nallapati, Zhai, and
  Zhou}]{Nallapati2017SummaRuNNer:Documents}
Ramesh Nallapati, Feifei Zhai, and Bowen Zhou. 2017.
\newblock {SummaRuNNer: A recurrent neural network based sequence model for
  extractive summarization of documents}.
\newblock In \emph{31st AAAI Conference on Artificial Intelligence, AAAI 2017},
  pages 3075--3081.

\bibitem[{Narayan et~al.(2018)Narayan, Cohen, and
  Lapata}]{narayan-etal-2018-dont}
Shashi Narayan, Shay~B. Cohen, and Mirella Lapata. 2018.
\newblock \href {https://doi.org/10.18653/v1/D18-1206} {Don{'}t give me the
  details, just the summary! topic-aware convolutional neural networks for
  extreme summarization}.
\newblock In \emph{Proceedings of the 2018 Conference on Empirical Methods in
  Natural Language Processing}, pages 1797--1807, Brussels, Belgium.
  Association for Computational Linguistics.

\bibitem[{Ruan et~al.(2022)Ruan, Ostendorff, and
  Rehm}]{ruan-etal-2022-histruct}
Qian Ruan, Malte Ostendorff, and Georg Rehm. 2022.
\newblock \href {https://doi.org/10.18653/v1/2022.findings-acl.102}
  {{H}i{S}truct+: Improving extractive text summarization with hierarchical
  structure information}.
\newblock In \emph{Findings of the Association for Computational Linguistics:
  ACL 2022}, pages 1292--1308, Dublin, Ireland. Association for Computational
  Linguistics.

\bibitem[{Wang et~al.(2020)Wang, Liu, Zheng, Qiu, and
  Huang}]{wang-etal-2020-heterogeneous}
Danqing Wang, Pengfei Liu, Yining Zheng, Xipeng Qiu, and Xuanjing Huang. 2020.
\newblock \href {https://doi.org/10.18653/v1/2020.acl-main.553} {Heterogeneous
  graph neural networks for extractive document summarization}.
\newblock In \emph{Proceedings of the 58th Annual Meeting of the Association
  for Computational Linguistics}, pages 6209--6219, Online. Association for
  Computational Linguistics.

\bibitem[{Wang et~al.(2018)Wang, Li, and Yang}]{wang-etal-2018-toward}
Yizhong Wang, Sujian Li, and Jingfeng Yang. 2018.
\newblock \href {https://doi.org/10.18653/v1/D18-1116} {Toward fast and
  accurate neural discourse segmentation}.
\newblock In \emph{Proceedings of the 2018 Conference on Empirical Methods in
  Natural Language Processing}, pages 962--967, Brussels, Belgium. Association
  for Computational Linguistics.

\bibitem[{Xiao et~al.(2020)Xiao, Huber, and Carenini}]{xiao-etal-2020-really}
Wen Xiao, Patrick Huber, and Giuseppe Carenini. 2020.
\newblock \href {https://doi.org/10.18653/v1/2020.codi-1.13} {Do we really need
  that many parameters in transformer for extractive summarization? discourse
  can help !}
\newblock In \emph{Proceedings of the First Workshop on Computational
  Approaches to Discourse}, pages 124--134, Online. Association for
  Computational Linguistics.

\bibitem[{Xu and Durrett(2019)}]{xu-durrett-2019-neural}
Jiacheng Xu and Greg Durrett. 2019.
\newblock \href {https://doi.org/10.18653/v1/D19-1324} {Neural extractive text
  summarization with syntactic compression}.
\newblock In \emph{Proceedings of the 2019 Conference on Empirical Methods in
  Natural Language Processing and the 9th International Joint Conference on
  Natural Language Processing (EMNLP-IJCNLP)}, pages 3292--3303, Hong Kong,
  China. Association for Computational Linguistics.

\bibitem[{Xu et~al.(2020)Xu, Gan, Cheng, and Liu}]{xu-etal-2020-discourse}
Jiacheng Xu, Zhe Gan, Yu~Cheng, and Jingjing Liu. 2020.
\newblock \href {https://doi.org/10.18653/v1/2020.acl-main.451}
  {Discourse-aware neural extractive text summarization}.
\newblock In \emph{Proceedings of the 58th Annual Meeting of the Association
  for Computational Linguistics}, pages 5021--5031, Online. Association for
  Computational Linguistics.

\bibitem[{Yoshida et~al.(2014)Yoshida, Suzuki, Hirao, and
  Nagata}]{yoshida-etal-2014-dependency}
Yasuhisa Yoshida, Jun Suzuki, Tsutomu Hirao, and Masaaki Nagata. 2014.
\newblock \href {https://doi.org/10.3115/v1/D14-1196} {Dependency-based
  discourse parser for single-document summarization}.
\newblock In \emph{Proceedings of the 2014 Conference on Empirical Methods in
  Natural Language Processing ({EMNLP})}, pages 1834--1839, Doha, Qatar.
  Association for Computational Linguistics.

\bibitem[{Zeldes et~al.(2019)Zeldes, Das, Maziero, Antonio, and
  Iruskieta}]{zeldes-etal-2019-disrpt}
Amir Zeldes, Debopam Das, Erick~Galani Maziero, Juliano Antonio, and Mikel
  Iruskieta. 2019.
\newblock \href {https://doi.org/10.18653/v1/W19-2713} {The {DISRPT} 2019
  shared task on elementary discourse unit segmentation and connective
  detection}.
\newblock In \emph{Proceedings of the Workshop on Discourse Relation Parsing
  and Treebanking 2019}, pages 97--104, Minneapolis, MN. Association for
  Computational Linguistics.

\bibitem[{Zhang et~al.(2019)Zhang, Wei, and Zhou}]{zhang-etal-2019-hibert}
Xingxing Zhang, Furu Wei, and Ming Zhou. 2019.
\newblock \href {https://doi.org/10.18653/v1/P19-1499} {{HIBERT}: Document
  level pre-training of hierarchical bidirectional transformers for document
  summarization}.
\newblock In \emph{Proceedings of the 57th Annual Meeting of the Association
  for Computational Linguistics}, pages 5059--5069, Florence, Italy.
  Association for Computational Linguistics.

\bibitem[{Zhong et~al.(2020)Zhong, Liu, Chen, Wang, Qiu, and
  Huang}]{zhong-etal-2020-extractive}
Ming Zhong, Pengfei Liu, Yiran Chen, Danqing Wang, Xipeng Qiu, and Xuanjing
  Huang. 2020.
\newblock \href {https://doi.org/10.18653/v1/2020.acl-main.552} {Extractive
  summarization as text matching}.
\newblock In \emph{Proceedings of the 58th Annual Meeting of the Association
  for Computational Linguistics}, pages 6197--6208, Online. Association for
  Computational Linguistics.

\end{thebibliography}
\bibliographystyle{acl_natbib}

\appendix

\section{Parameters for Oracle Summaries}
\label{appendix: os length}
Table \ref{tab:oracle analysis params} presents parameters for oracle summaries.

\begin{table}[ht]
    \centering
    \begin{tabular}{lcc}
    \hline
        Dataset & \# Sentences & \# EDUs \\
        \hline
        CNN/DM & 5 & 8 \\
        XSum & 5 & 8 \\
        Reddit & 5 & 8 \\
        WikiHow & 5 & 8 \\
        Multi-News & 15 & 30 \\
        \hline
    \end{tabular}
    \caption{Maximum number of textual segments allowed to be extracted in oracle summaries.}
    \label{tab:oracle analysis params}
\end{table}

\section{Breakdown Comparison on ROUGE scores}
\label{appendix: breakdown rouge}

Table \ref{appendix:breakdown oracle analysis} presents the breakdown ROUGE scores of other four datasets.
\begin{table}[ht]
    \centering
    \begin{tabular}{l c c c c}
        \hline
         & \multicolumn{2}{c}{Sentence} & \multicolumn{2}{c}{EDU} \\
        \cmidrule(l){2-3}\cmidrule(l){4-5}
        Metric & recall & precision & recall & precision \\
        \hline
        \multicolumn{5}{c}{\textbf{XSum}} \\
        \hline
        R-1 & 40.18 & 25.77 & 40.16 & 36.54  \\
        R-2 & 11.70 & 7.95 & 12.86 & 12.26  \\
        R-L & 30.68 & 19.79 & 34.31 & 31.44  \\
        \hline
        \multicolumn{5}{c}{\textbf{WikiHow}} \\
        \hline
        R-1 & 45.28 & 36.90 & 44.41 & 49.25  \\
        R-2 & 16.45 & 13.44 & 18.01 & 19.99  \\
        R-L & 41.96 & 34.17 & 42.71 & 47.29  \\
        \hline
        \multicolumn{5}{c}{\textbf{Reddit}} \\
        \hline
        R-1 & 44.70 & 26.71 & 45.40 & 40.39  \\
        R-2 & 15.63 & 10.02 & 17.62 & 16.48  \\
        R-L & 35.86 & 21.56 & 40.19 & 35.75  \\
        \hline
        \multicolumn{5}{c}{\textbf{Multi-News}} \\
        \hline
        R-1 & 45.09 & 58.87 & 42.45 & 68.35  \\
        R-2 & 19.96 & 26.72 & 19.86 & 31.79  \\
        R-L & 40.77 & 53.44 & 40.24 & 64.86  \\
        \hline
    \end{tabular}
    \caption{Breakdown ROUGE scores of sentence/EDU-level oracle summaries on XSum, WikiHow, Reddit, and Multi-News training datasets.}
    \label{appendix:breakdown oracle analysis}
\end{table}

\section{Statistics of Datasets}
\label{appendix:datatset stats}
Table \ref{table:datatset stats} presents the statistics of the five datasets.

\begin{table}[ht]
    \centering
    \resizebox{\columnwidth}{!}{%
    \begin{tabular}{lcccc}
    \hline
        Dataset & \# word & \# EDU & \# sent. & \# EDU/sent. \\
        \hline
        CNN/DM & 733.98 & 94.25 & 36.23 & 2.67 \\
        XSum & 431.12 & 52.02 & 19.76 & 2.63 \\
        Reddit & 443.46 &  65.28 & 23.44 & 3.01 \\
        WikiHow & 581.15 &  75.72 & 29.42 & 2.58 \\
        Multi-News & 503.33 & 58.33 & 18.13 & 3.35 \\
        \hline
    \end{tabular}%
    }
    \caption{Statistics of datasets. \#word, \#EDU and \#sent. refer to the average number of words, EDUs and sentences, respectively, of documents in the dataset. \#EDU/sent. refers to the average number of EDUs per sentence.}
    \label{table:datatset stats}
\end{table}

\section{Greedy Selection Algorithm}
\label{appendix: greedy algo}
Algorithm \ref{algo: greedy} presents the pseudo-code of the algorithm of selecting salient textual segments, which is used to generate oracle summary and ground truth labels.

\begin{algorithm}
\caption{Greedy Selection Algorithm}
\label{algo: greedy}
\begin{algorithmic}[1]
\Require{$Doc, Ref, k$} \Comment{$k$: \# of selections}
\Ensure{$sel\_idx$} \Comment{selected indices}
\State {$sel\_idx$ $\gets$ [ ]} \Comment{empty list}
\State {$C \gets $ [ ]} \Comment{candidate: empty list}
\While{$k \geq 0$}
\State {end $\gets$ TRUE }
\For{$i \gets 0$ to $len(Doc)$}
  \State {$tmp\_C$ $\gets$ $ C + [Doc_i]$}
  \State {$score$ $\gets$ $ROUGE(tmp\_C, Ref)$}
  \If {$score$ increases}
    \State {$sel\_idx$ $\gets$ $sel\_idx + [i]$}
    \State {$C \gets tmp\_C$}
    \State {$k \gets k-1$}
    \State {end $\gets$ FALSE}
    \State break
\EndIf
\EndFor
\If {end}
  \State break
\EndIf
\EndWhile
\State \textbf{return} $sel\_idx$
\end{algorithmic}
\end{algorithm}

\section{Supplementary Experimental Settings and Results}
\label{appendix: more settings}
Table \ref{table: model supplement} and Table \ref{table: result supplement} present detailed experimental settings and results, respectively.
\begin{table}[ht]
    \centering
    \resizebox{\columnwidth}{!}
    {%
    \begin{tabular}{lccc}
        \hline
        \multicolumn{3}{c}{\textbf{Model Statistics}} \\
        \hline
        model & \#params & runtime per epoch  \\
        \hline
        $\textsc{EDU-VL}_\textsc{RoBERTa}$ & 147M & 1h 20min \\
        $\textsc{EDU-VL}_\textsc{BART}$ & 161M & 1h 30min \\
        \hline
        \multicolumn{3}{c}{\textbf{Pre-processing Setting}} \\
        \hline
        dataset & \#min & \#max \\
        \hline
        CNN/DM & 6 & 10 \\
        XSum & 3 & 7 \\
        Reddit & 4 & 8 \\
        WikiHow & 6 & 10 \\
        Multi-News & 27 & 31 \\
        \hline
    \end{tabular} %
    }
    \caption{Supplementary information of experimental settings. \#params refers to the total number of trainable parameters in the model (here both versions are calculated with 4 MTLs). \#min and \#max refer to the range of lengths ($k$ values in the top-$k$ strategy) of candidate summaries generated by the model, respectively.}
    \label{table: model supplement}
\end{table}

\begin{table}[ht]
    \centering
    \begin{tabular}{lccc}
        \hline
        Model & R-1 & R-2 & R-L \\
        \hline
        \multicolumn{4}{c}{\textbf{CNN/DM}} \\
        \hline
        $\textsc{EDU-VL}_\textsc{RoBERTa}$ & 45.45 & 22.10 & 43.23 \\
        $\textsc{EDU-VL}_\textsc{BART}$ & 45.29 & 22.08 & 41.11 \\
        \hline
        \multicolumn{4}{c}{\textbf{XSum}} \\
        \hline
        $\textsc{EDU-VL}_\textsc{RoBERTa}$ & 26.58 & 5.83 & 22.34 \\
        $\textsc{EDU-VL}_\textsc{BART}$ & 26.66 & 5.97 & 22.51 \\
        \hline
        \multicolumn{4}{c}{\textbf{Reddit}} \\
        \hline
        $\textsc{EDU-VL}_\textsc{RoBERTa}$ & 28.20 & 7.84 & 23.58 \\
        $\textsc{EDU-VL}_\textsc{BART}$ & 28.40 & 7.81 & 23.89 \\
        \hline
        \multicolumn{4}{c}{\textbf{WikiHow}} \\
        \hline
        $\textsc{EDU-VL}_\textsc{RoBERTa}$ & 33.90 & 10.19 & 32.53 \\
        $\textsc{EDU-VL}_\textsc{BART}$ & 33.95 & 10.31 & 32.59 \\
        \hline
        \multicolumn{4}{c}{\textbf{Multi-News}} \\
        \hline
        $\textsc{EDU-VL}_\textsc{RoBERTa}$ & 46.58 & 17.00 & 44.14 \\
        $\textsc{EDU-VL}_\textsc{BART}$ & 47.29 & 17.49 & 44.82 \\
        \hline
    \end{tabular} 
    \caption{Experimental results of ROUGE F1-scores on the corresponding validation datasets.}
    \label{table: result supplement}
\end{table}

\end{document}